\documentclass{article}

    \usepackage[preprint]{neurips_2025}

\usepackage[utf8]{inputenc} 
\usepackage[T1]{fontenc}    
\usepackage{hyperref}       
\usepackage{url}            
\usepackage{booktabs}       
\usepackage{amsfonts}       
\usepackage{nicefrac}       
\usepackage{microtype}      
\usepackage{xcolor}         
\usepackage{amsmath}
\usepackage{graphicx}       
\usepackage{cleveref}
\usepackage{multirow}
\usepackage{subcaption}
\usepackage{enumitem}
\setlist[itemize]{leftmargin=0.5cm}

\usepackage[textwidth=2.5cm]{todonotes}

\title{Your LLM Knows the Future: \\ Uncovering Its Multi-Token Prediction Potential}

\author{
Mohammad Samragh$^{*\dagger}$
\And
Arnav Kundu$^{*}$
\And
David Harrison$^{*}$
\AND
Kumari Nishu
\And Devang Naik
\And Minsik Cho 
\And Mehrdad Farajtabar$^{\dagger}$ \AND
\texttt{Apple}
}

\begin{document}
\maketitle
\vspace{-0.5cm}
\let\thefootnote\relax\footnotetext{$^{*}$ Equal Contributor Authors. 
\par $\ \ \ ^{\dagger}$ Corresponding authors. Emails: m\_samraghrazlighi@apple.com, farajtabar@apple.com.}

\vspace{-1mm}
\begin{abstract}



\vspace{-1mm}
Autoregressive language models are constrained by their inherently sequential nature, generating one token at a time. This paradigm limits inference speed and parallelism, especially during later stages of generation when the direction and semantics of text are relatively certain. 
In this work, we propose a novel framework that leverages the inherent knowledge of vanilla autoregressive language models about future tokens, combining techniques to realize this potential and enable simultaneous prediction of multiple subsequent tokens.
Our approach introduces several key innovations: (1) a masked-input formulation where multiple future tokens are jointly predicted from a common prefix; (2) a gated LoRA formulation that preserves the original LLM's functionality, while equipping it for multi-token prediction; (3) a lightweight, learnable sampler module that generates coherent sequences from the predicted future tokens; (4) a set of auxiliary training losses, including a consistency loss, to enhance the coherence and accuracy of jointly generated tokens; and (5) a speculative generation strategy that expands tokens quadratically in the future while maintaining high fidelity. Our method achieves significant speedups through supervised fine-tuning on pretrained models. For example, it generates code and math nearly $5\times$ faster, and improves general chat and knowledge tasks by almost $2.5\times$. These gains come without any loss in quality. 
\vspace{-2mm}
\end{abstract}

\vspace{-1mm}
\vspace{-2mm}
\section{Introduction}
\vspace{-2mm}

Recent advances in language models are largely driven by the availability of large-scale text data and the effectiveness of autoregressive training, where each token serves as the target for its preceding context~\citep{radford2018improving, brown2020language}. This framework eliminates the need for explicit labels and has made autoregressive models the dominant paradigm due to the advantages that it offers at training time. However, at inference time, autoregressive generation is inherently sequential and computationally expensive, as it requires full model execution at each decoding step. In contrast, humans often formulate thoughts at the sentence level before articulating them word by word. Motivated by this, we investigate whether a similar strategy can improve the efficiency of text generation by language models.

We begin by addressing a fundamental question: can a language model generate multiple tokens in a single inference step? Encouragingly, the answer is yes. Existing works on speculative decoding have already explored this direction to accelerate generation~\citep{cai2024medusa,bhendawade2024speculative,cheng2024recurrent}. Speculative decoding methods leverage a draft model to generate multiple tokens, followed by a verifier that checks their consistency with standard autoregressive outputs. While this approach offers speedup, it still fundamentally relies on autoregressive generation. In this work, we ask a deeper question: can we train genuinely non-autoregressive language models? 

We approach this question by designing fully non-autoregressive training algorithms, such as diffusion-based language models~\citep{nie2025large,gong2025diffucoder}. However, these methods often require building entirely new modeling and training pipelines. The next question we ask ourselves is, can we adapt the existing autoregressive training and inference setup with minimal changes? Our goal is to enable efficient multi-token generation while retaining the core benefits of autoregressive models.

\begin{figure}
    \centering
    \vspace{-2mm}
\includegraphics[width=0.90\linewidth]{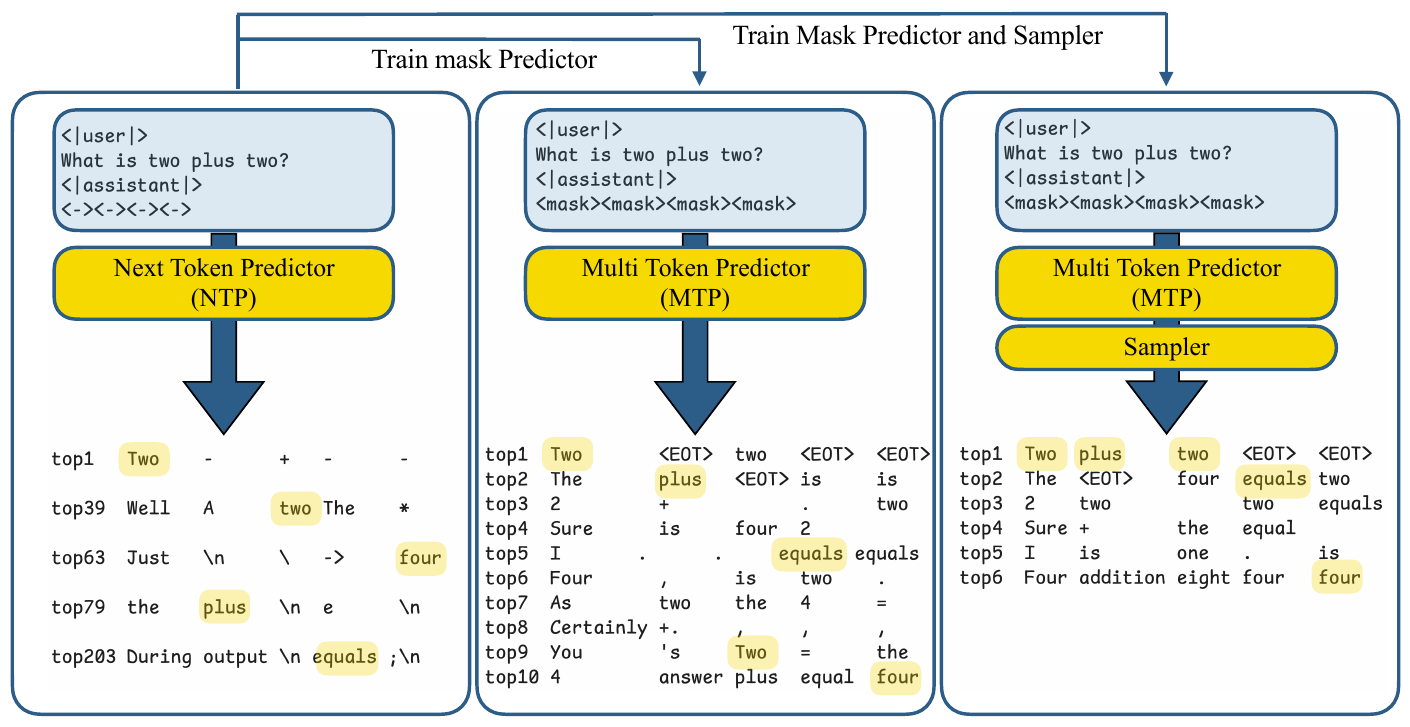}
    \vspace{-2mm}
    \caption{Autoregressive models implicitly anticipate future tokens. (Left): A pretrained model queried with a prompt plus redundant (-) tokens ranks the correct next token within the top-200. (Middle): Finetuning with \textit{<mask>} tokens improves structure, pushing correct predictions into the top-10. (Right): Training a \textit{sampler head} further refines future token prediction.}
    \label{fig:motivation}
    \vspace{-3mm}
\end{figure}

To motivate this, we first observe that autoregressive models already encode some information about future tokens, even though they are not explicitly trained for it. For example, given the prompt "what is two plus two?", a pretrained model will generate "two plus two equals four" in standard autoregressive decoding. To test the model's awareness of future tokens, we append placeholder tokens (shown by \texttt{<->} in the figure) to the prompt and examine the output logits, as illustrated in the left panel of Figure~\ref{fig:motivation}. Surprisingly, the correct sequence of future tokens appears within the top 200 logits, suggesting the model has implicit knowledge of upcoming tokens.

Building on this insight, we investigate whether the model can be guided to realize this potential and better structure its predictions of future tokens. We introduce mask tokens at the end of the prompt and train the model to predict them directly. As shown in the middle panel of Figure~\ref{fig:motivation}, the finetuned model can surface correct tokens in the top-10 logits. Finally, to produce coherent multi-token outputs, we incorporate a lightweight sampling module: a two-layer perceptron that conditions each predicted token on previously sampled ones, illustrated in the right panel of Figure~\ref{fig:motivation}. 

\begin{figure}[!h]
    \centering
    \vspace{-0.25cm}
\includegraphics[width=0.47\linewidth]{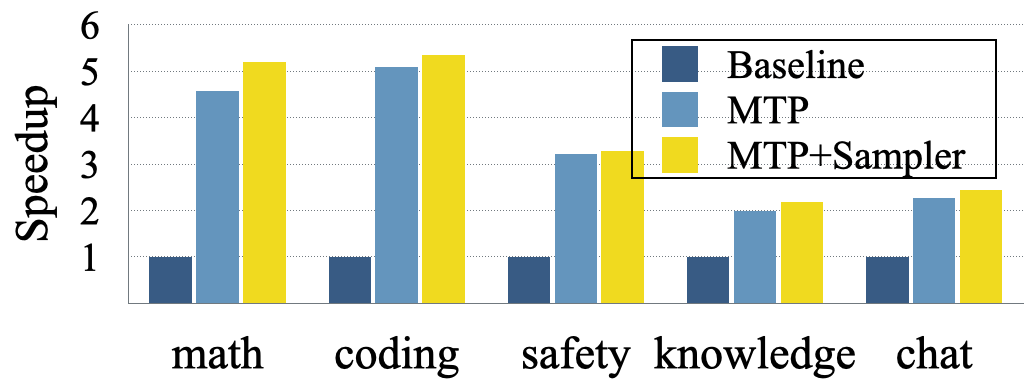}
    \vspace{-0.2cm}
    \caption{Speedup achieved after training gated LoRA and sampler head with supervised finetuning.}
    \label{fig:per_task}
    \vspace{-0.3cm}
\end{figure}

Our work follows in the tradition of multi-token prediction (MTP) ~\citep{Gloeckle2024BetterF} and speculative decoding ~\citep{Leviathan2022FastIF}. Most related to our work, using similar additional tokens for speculative decoding and MTP, are ~\citep{Gerontopoulos2025MultiTokenPN,Chen2024HardwareAwarePP,Liu2024SDSATAL,Xiao2024ParallelSpecPD}. However, our model is trained to fill in mask tokens with future tokens. It uses its full depth and capacity to infer these tokens, leveraging the entire context of the sequence. This provides a significant advantage over existing multi-token prediction methods. Additionally, unlike prior work, our method guarantees no degradation in generation quality, thanks to a simple yet effective technique we call gated LoRA adaptation. 
With relatively lightweight supervised fine-tuning, we achieve speedups compared to baseline retrogressive model ($1\times$) as shown in Figure~\ref{fig:per_task}, where the Tulu3-8B model~\citep{lambert2024t} is fine-tuned to predict 8 additional tokens.

\section{Methodology}

To enable multi-token generation with minimal retraining, we introduce special tokens called \emph{masks}. Let \( X = [x_1, \dots, x_n] \) denote a sequence of tokens. The idea is to append \( k \) unique mask tokens to the end of the sequence, and form an augmented sequence \( X_m = [x_1, \dots, x_n, m_1, \dots, m_k] \). The representations of $[m_1, \dots, m_k]$ are generated as random vectors and added to the embedding table of the model. Throughout this work, we refer to the model’s standard next-token predictions as \textbf{NTP}, and predictions for the mask tokens as \textbf{MTP}.

\begin{figure}
    \centering
    \includegraphics[width=\linewidth]{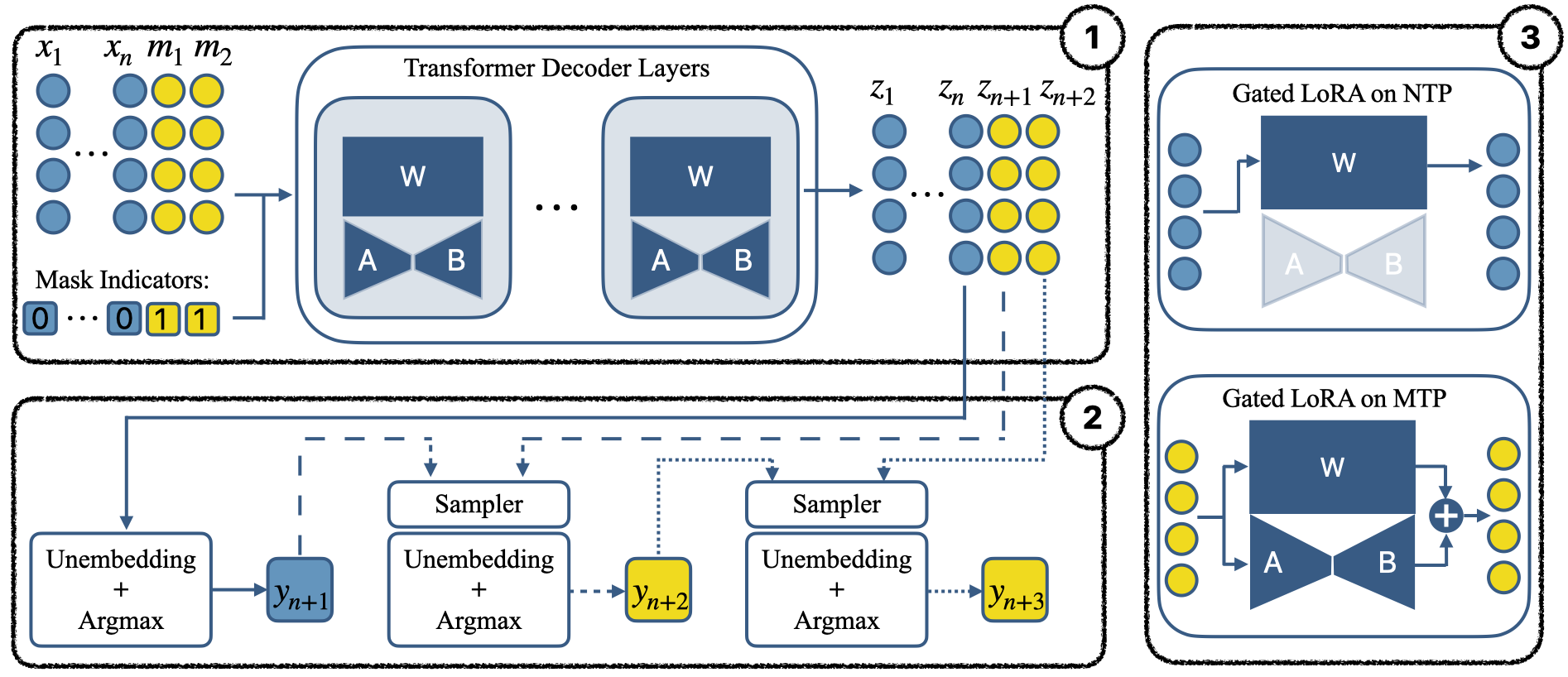}
    \caption{Components of our MTP model. Box-1 (top-left) shows the autoregressive model augmented with gated LoRA parameters. Box-2 (bottom left) illustrates the sampler head. Box-3 (right) presents the block diagram of the gated LoRA module.}
    \label{fig:architecture}
\end{figure}

The overall architecture of our MTP model is illustrated in Figure~\ref{fig:architecture}, where the \textbf{inference stage} is shown for a model that was finetuned with $k=2$ masks. In box-1 (top left), \( X_m \) is fed into the decoder and the network produces latent representations \( [z_1, \dots, z_n] \) for the NTP tokens, and \( [z_{n+1}, \dots, z_{n+k}] \) for the MTP tokens.

Box-2 (bottom-left) shows the \emph{sampler head}. The first (NTP) token is generated autoregressively using the standard unembedding layer, i.e., \( y_{n+1} \) is predicted from \( z_n \). The remaining (MTP) tokens are produced sequentially by the sampler module. At each step, the sampler generates \( y_{n+1+k} \) conditioned on both \( z_{n+k+1} \) and the previously sampled token \( y_{n+k} \), ensuring that each sampled token incorporates both the model’s latent representation and prior sampled tokens.

To enable fine-tuning while preserving the pretrained model’s behavior, we augment the decoder layers with \emph{gated LoRA} modules. During fine-tuning, only the LoRA parameters and sampler head parameters are updated, while the original decoder weights remain frozen. The gated LoRA module ensures that fine-tuning does not affect NTP token behavior by applying distinct functional paths for NTP and MTP tokens, as illustrated in box-3 (right). This distinction is achieved using binary masks, which are passed as additional inputs to the decoder layers. 

The explanations address the inference flow of our multi-token generation. We next describe how such a network can be trained.

\subsection{Training with Mask Tokens}
\begin{figure}
    \centering
    \includegraphics[width=0.86\linewidth]{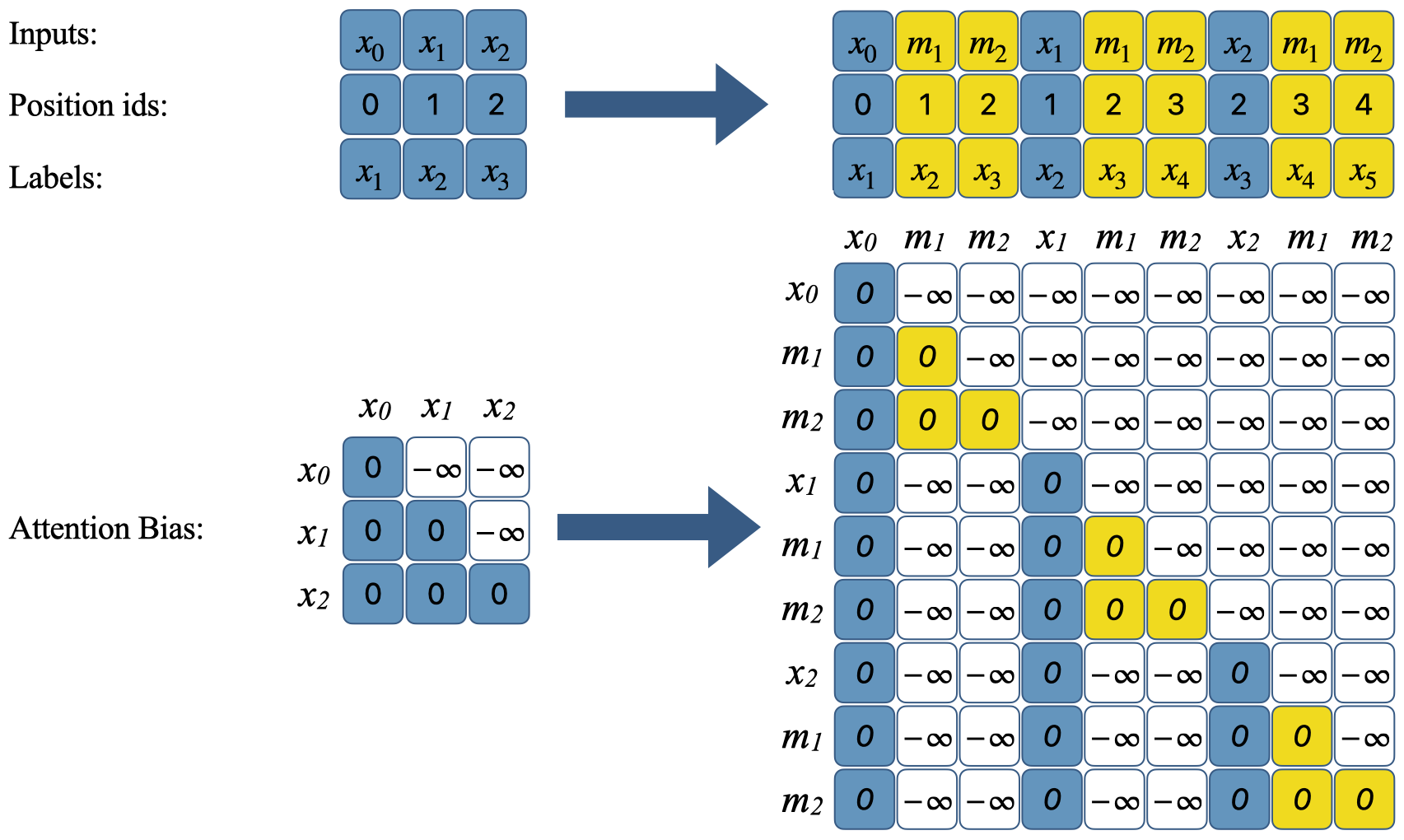}
    \caption{Converting regular inputs to masked inputs for predicting 2 extra tokens during training. NTP and MTP tokens are shown in blue and yellow, respectively. NTP tokens attend only to previous NTP tokens, which guarantees that the model's output for NTP tokens does not change. MTP tokens attend to previous NTP tokens and MTP tokens of the same block, but not to earlier MTP blocks.
}
    \label{fig:masked_inputs}
\end{figure}

Figure~\ref{fig:architecture} illustrated how the appended mask tokens are used \textbf{during inference}. During training, however, we must insert $k$ masks after NTP tokens in different positions of the input sequence. To enable an efficient implementation, we modify the input tokens, position IDs, labels, and attention biases, as shown in Figure~\ref{fig:masked_inputs}. 
We highlight several notes about the input modifications:
\begin{itemize}
\item For an original sequence of length $n$, the modified input effectively simulates $n$ separate prompts, where the $i$-th prompt includes the first $i$ tokens followed by $k$ masks. All such $n$ queries are processed in parallel within a single model invocation, providing significant training speedup.
\item The modifications preserve the model’s predictions on the original NTP tokens, ensuring consistent behavior before and after applying the changes. 
\item If an NTP token does not require loss calculation (e.g., "user prompt" tokens during supervised fine-tuning), no masks are inserted after it.
\end{itemize}

\subsection{Preserving Functionality with Gated LoRA}
\label{sub:gated_lora}
Multi-token prediction (MTP) can be applied at various stages of language model development. In this work, we focus on applying it during the \emph{fine-tuning} stage to reduce training costs. However, fine-tuning introduces the risk of degrading the model’s original performance, particularly on next-token prediction (NTP) tasks.

To address this, we introduce a \emph{gated version of LoRA}~\citep{hu2022lora}, where low-rank adapters are added in parallel to the model’s existing linear layers, but are only activated for MTP tokens. Specifically, in a standard LoRA layer, given an input \( x_t \in \mathbb{R}^d \) at position $t$, the output is:

\[
y_t = W\cdot x_t + A\cdot B \cdot x_t,
\]

where \( W\cdot x_t \) is the original model’s linear transformation, and \( A \cdot B \cdot x_t \) is the LoRA adaptation, with \( A \in \mathbb{R}^{d \times r} \) and \( B \in \mathbb{R}^{r \times d} \). In our \emph{gated LoRA}, the output becomes:

\[
y = W \cdot x_t + I(t) [A \cdot B \cdot x_t],
\]

where \( I(t) \in \{0, 1\} \) is an indicator function that activates the LoRA path only for MTP tokens (\( I(t) = 1 \)) and disables it for NTP tokens (\( I(t) = 0 \)). Note that the gate value $I(t)$ is deterministically known for any $t$, as we know whether the token at index $t$ was a mask token or a regular token.

This gating ensures that the outputs for NTP tokens remain unchanged from the base model, preserving its original behavior. When combined with \emph{speculative decoding} (see Section~\ref{sec:specD}), this strategy allows the model to generate multiple tokens per step while maintaining consistency with standard autoregressive generation.


\subsection{Sampling from Predicted Tokens}
A language model trained with mask tokens can predict future token distributions, as indicated in the middle panel of Figure~\ref{fig:motivation}. However, the coherence of the generated sequence depends on how individual tokens are selected. To enhance sequence coherence, we propose to train a sampler head depicted in the right panel of Figure~\ref{fig:motivation}. Let $z_n \in \mathbb{R}^d$ denote the hidden representation of a masked token at position $n$, as produced by the decoder layers of the language model. In standard decoding, this hidden vector is transformed into token logits via the un-embedding matrix $W \in \mathbb{R}^{V \times d}$, yielding $p_n = W \cdot z_n$, where $V$ is the vocabulary size. This produces a distribution over tokens conditioned solely on $z_n$.

In contrast, our sampling strategy conditions $p_n$ on both $z_n$ and the previously sampled token $y_{n-1}$. To achieve this, we retrieve the embedding vector $E_{y_{n-1}} \in \mathbb{R}^d$ for token $y_{n-1}$ from the model's input embedding table $E \in \mathbb{R}^{V \times d}$, and concatenate it with $z_n$. The concatenated vector $[E_{y_{n-1}}; z_n] \in \mathbb{R}^{2d}$ is then passed through a two-layer MLP to produce a transformed feature vector. The final logits are computed as: 

\[
p_n = W \cdot \text{MLP}([E_{y_{n-1}}; z_n]),
\]

where $MLP(\cdot)$ is a perceptron with two internal blocks, and $W$ is the unembedding layer. Each internal block of the perceptron is a stack of a Linear layer followed by \texttt{SiLU} activation and \texttt{LayerNorm} operation. This formulation enables token predictions that are explicitly conditioned on both the current context and the preceding generated token, improving coherence in multi-token generation. 


\subsection{Speculative Decoding to Verify Predicted Tokens}\label{sec:specD}

To leverage multi-token prediction, a straightforward approach is to generate 
$k+1$ tokens at each generation step instead of just one. However, this naive method can lead to incorrect outputs, especially for longer sequences (i.e., large $k$). To mitigate this, it is necessary to verify whether the predicted future tokens are valid, specifically, whether the 
$k+1$ tokens generated in a single step match what the model would have produced over $k+1$ standard auto-regressive steps.

To address this challenge, we employ an speculative decoding scheme. Let \( X^t = [x_1, \dots, x_n] \) represent the sequence of tokens generated up to iteration \( t \). The superscript \( t \) denotes the current generation step, while the subscript indicates the token's position in the sequence. To generate future tokens, we append \( k \) unique mask tokens to the input, resulting in the sequence:
\[
X^t_m = 
\underbrace{[x_1, \dots, x_n]}_{\text{verified tokens}} 
+ 
\underbrace{[m_1, \dots, m_k]}_{\text{augmented masks}}.
\]

When this input \( X^t_m \) is fed into the model, it produces \(k + 1 \) new tokens:
\[
\underbrace{[x^t_{n+1}]}_{\text{verified}} + \underbrace{[x^t_{n+2}, \dots, x^t_{n+k+1}]}_{\text{speculated}}.
\]
Here, \( x^t_{n+1} \) represents the next token the model would produce in a standard auto-regressive manner. Since this token is generated exactly as it would be in regular decoding, we consider it \emph{verified} and drop the superscript \( t \), denoting it as \( x_{n+1} \). The remaining \( k \) tokens, \( x^t_{n+2}, \dots, x^t_{n+k+1} \), are \emph{speculative predictions} generated in step \( t \), and their correctness must be validated. To emphasize that they were speculated during step \( t \), we retain the superscript \( t \) in the notation.

At this point, we have \( k \) speculative tokens that we would like to verify and potentially add to the final sequence. To do so, we propose two decoding strategies: \textbf{Linear Decoding} and \textbf{Quadratic Decoding}. We begin by describing the linear method.

\subsubsection{Linear Decoding}

In linear decoding, at the next generation step \( t+1 \), we construct a new input sequence by concatenating the verified portion of the sequence, including \( x_{n+1} \), the \( k \) speculative tokens from step \( t \), and
\( k \) new mask tokens for further prediction:

\[
X^{t+1}_m = 
\underbrace{[x_1, \dots, x_n, x_{n+1}]}_{\text{verified}} 
+ 
\underbrace{[x^t_{n+2}, \dots, x^t_{n+k+1}]}_{\text{speculated}} 
+ 
\underbrace{[m_1, \dots, m_k]}_{\text{augmented masks}}.
\]

Feeding this input to the model yields a new set of predictions:

\[
\underbrace{[x^{t+1}_{n+2}]}_{\text{verified}} 
+ \underbrace{[x^{t+1}_{n+3}, \dots, x^{t+1}_{n+k+1}]}_{\text{to be verified}} 
+ \underbrace{[x^{t+1}_{n+k+2}, \dots, x^{t+1}_{n+2k+1}]}_{\text{speculated}}.
\]

As before, the first token \( x^{t+1}_{n+2} \) is generated autoregressively and automatically considered verified. We then compare \( x^{t+1}_{n+2} \) with \( x^t_{n+2} \) from the previous generation step. If they match, the next token \( x^{t+1}_{n+3} \) is similarly validated. This verification process continues sequentially for tokens \( x^t_{n+3}, \dots, x^t_{n+k} \). Depending on the outcome of these comparisons, two scenarios can occur:

\begin{enumerate}
    \item \textbf{Full verification:} All comparisons succeed, i.e., \( x_{n+j}^{t+1} = x_{n+j}^t \) for all \( 2 \leq j \leq k \). In this case, the verified sequence extends to \( [x_1, \dots, x_{n+k+1}] \). The set of  speculative tokens \( [x^{t+1}_{n+k+2}, \dots, x^{t+1}_{n+2k+1}] \) can be used in the next generation step.
    \item \textbf{Partial verification:} Full verification fails, i.e., \( x_{n+j}^{t+1} \neq x_{n+j}^t \) at some position \( 2 \leq j \leq k \). In this case, the verified sequence includes only \( [x_1, \dots, x_{n+j}] \), and no speculative tokens are available for the next generation step.
\end{enumerate}

In summary, speculative tokens are only available for the next step if full verification succeeds (case 1). This requirement can limit the potential speedup of multi-token prediction. To address this limitation, we propose a more robust strategy, called \emph{quadratic decoding}, which guarantees that exactly \( k \) tokens are always available for verification at each generation step.

\subsubsection{Quadratic Decoding}

Unlike linear decoding, which appends augmented mask tokens only at the end of the sequence, \emph{quadratic decoding} interleaves mask tokens within the speculative tokens. Specifically, the input at generation step \( t+1 \) is constructed as:

\[
X^{t+1}_m = 
\underbrace{[x_1, \dots, x_n, x_{n+1}]}_{\text{verified}} 
+ 
\underbrace{[x^t_{n+2}, m_1, \dots, m_k]}_{\text{speculated \& augmented}} 
+ 
\dots 
+ 
\underbrace{[x^t_{n+k+1}, m_1, \dots, m_k]}_{\text{speculated \& augmented}}.
\]

The total number of inserted masks in this approach is \( k^2 \), which gives the method its name: \emph{quadratic decoding}. With this interleaved design, even if the verification of speculative tokens fails at some index, the sequence will still contain newly inserted mask tokens for further prediction in the next step. This ensures that every generation step produces exactly \( k \) new speculative tokens to be verified, maintaining consistent decoding progress. Importantly, interleaving masks in this manner does not interfere with future predictions, thanks to a carefully constructed attention bias (see Figure~\ref{fig:masked_inputs}). This bias restricts speculative tokens from attending to the speculative extensions of previously generated speculative tokens. This pattern of attention control has also been referred to as \emph{Tree Attention} in prior work~\cite{Miao2023SpecInferAL, Spector2023AcceleratingLI}.

Quadratic decoding guarantees an acceptance rate at least as high as linear decoding. Unlike linear decoding, where adding more speculative tokens can lower acceptance by making full verification harder, quadratic decoding ensures that increasing speculative tokens does not reduce acceptance. The trade-off is higher parallel computation: linear decoding processes sequences of length \(n + k\), while quadratic decoding handles sequences of length \(n + k^2\). However, since \(k \ll n\) in practice, the additional cost of quadratic decoding is negligible.

\subsection{Training Loss}

\noindent{\bf Cross-entropy loss:}
Let $(x_t, y_t)$ represent a pair consisting of an input token and its corresponding label at position $t$. The input token $x_t$ can be either an NTP or an MTP token, and the label $y_t$ can refer to either the immediate next token or a future token several steps ahead. For a vocabulary of size $v$, the model produces two output probability distributions: $p_t^b \in \mathbb{R}^v$ from the base unembedding layer, and $p_t^s \in \mathbb{R}^v$ from the sampler head.

We compute the cross-entropy loss for each of these distributions as follows:
\[
\mathcal{L}_t^b = -\log p_t^b(y_t), \quad \mathcal{L}_t^s = -\log p_t^s(y_t),
\]
where $p_t^b(y_t)$ and $p_t^s(y_t)$ denote the probabilities assigned to the correct label $y_t$ by the base and sampler heads, respectively. We refer to $\mathcal{L}_t^b$ and $\mathcal{L}_t^s$ as the base and sampler cross-entropy losses.

\begin{figure}
    \centering
    \includegraphics[width=0.5\linewidth]{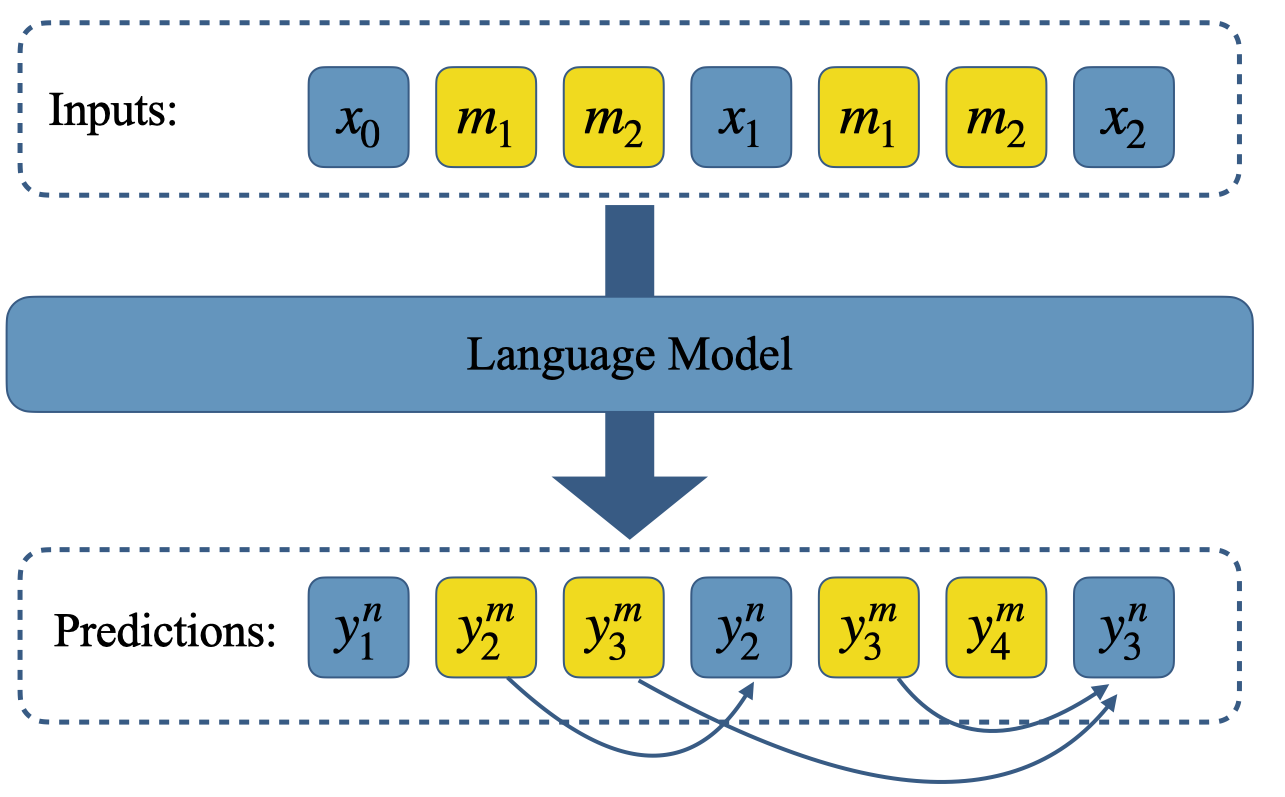}
    \caption{Illustration of the consistency loss. The goal is to minimize the distance between masked token predictions and the corresponding next-token predictions for each token. The arrows indicate which masked token predictions should be matched with which next-token predictions.}
    \label{fig:lcm}
\end{figure}

\noindent{\bf Latent consistency loss:} As discussed in Section~\ref{sec:specD}, the speedup gained by multi-token prediction depends on how often MTP predictions align with the autoregressive NTP outputs. To enhance this alignment—and thus improve speedup—we introduce a loss function that encourages consistency between MTP and NTP tokens. As shown in Figure~\ref{fig:lcm}, the goal is for the  MTP label prediction $(y^m_2 \mid x_0)$ to closely match the autoregressive representation $(y^n_2 \mid x_0, x_1)$, thereby ensuring that the masked token preserves the same output distribution as standard next-token prediction.

To enforce this property, we introduce \textit{latent consistency loss} (LCM). 
In designing this loss function, we were inspired the line of work in knowledge distillation~\citep{Luo2023LatentCM}.
Formally, let $z_t$ denote the hidden representation of the NTP token at index $t$, derived at the last decoder layer of the transformer architecture. For each such $z_t$, there may be up to $k$ corresponding MTP tokens. These tokens appear at earlier sequence positions $t' < t$ and have hidden representations that should ideally match $z_t$. We denote the set of these MTP representations as $S(z_t)$, where $|S(z_t)| \leq k$.

The goal is to bring each element in $S(z_t)$ as close as possible to the anchor $z_t$. To this end, we define the Latent Consistency Matching (LCM) loss as:

\begin{equation}
    \mathcal{L}_t^{\text{lcm}} = \frac{1}{|S(z_t)|} \sum_{z \in S(z_t)} \left( z_t - z \right)^2
\end{equation}

We make several important remarks: 
\begin{itemize}
    \item $\mathcal{L}_t^{\text{lcm}}$ is only defined when the index $t$ corresponds to an NTP token, i.e., a token used to generate the model’s output in an autoregressive fashion; 
    \item $z_t$ is detached from the computation graph for this loss, meaning gradients do not propagate through it. This ensures that only the MTP representations are encouraged to align with the anchor $z_t$, while the anchor itself remains unaffected by the loss.
    \item With the gated LoRA mechanism introduced in Section~\ref{sub:gated_lora}, the LCM loss acts as a form of self-distillation. This is because the NTP vector $z_t$ remains identical to that of the original autoregressive model.
\end{itemize}

\noindent{\bf Overall training loss:} The overall training loss for each input sample is computed as the average of loss values over specific token positions indexed by \( t \). Formally, the loss is defined as:
\[
\mathcal{L} = \underset{t \in T_{\text{ntp}} \cup T_{\text{mtp}}}{\mathbb{E}} \left[ \mathcal{L}^b_t + \mathcal{L}_t^s \right] + \underset{t \in T_{\text{ntp}}}{\mathbb{E}} \left[ \mathcal{L}^\text{lcm}_t\right].
\]
In this expression, \(\mathcal{L}^b_t\), \(\mathcal{L}^s_t\), and \(\mathcal{L}^\text{lcm}_t\) denote the base cross-entropy loss, the sampler cross-entropy loss, and the LCM loss at position \( t \), respectively, as introduced earlier in this section. The symbol \(\mathbb{E}\) represents the expectation (or average), and the set below it specifies the token indices over which this expectation is computed. The sets \(T_{\text{ntp}}\) and \(T_{\text{mtp}}\) represent NTP and MTP tokens, respectively.

\section{Experiments}

We perform our experiments on the Tulu3-8B SFT model~\citep{lambert2024t}. The model is part of the LLaMA-3 family and was supervised fine-tuned using the Tulu3 dataset. We chose Tulu3 because it is open-source and provides both the model weights and the full training dataset. The dataset spans nearly 1 million examples across diverse domains, including math, code, conversation, and scientific benchmarks. This breadth makes Tulu3 a strong candidate for evaluating our method’s effectiveness across multiple task types.

For experiments, we train a model with $k=8$ mask tokens, resulting in a model that can generate up to $k+1=9$ tokens at a time. Our masked token predictor can be trained at various stages of language model training. To reduce GPU usage and overall training costs, we adopt a fine-tuning setup with a limited number of training steps. Specifically, we freeze the main model and augment it with Gated-LoRA layers of rank 128. We also utilize a 2-layer MLP block to serve as the sampler module. Only the LoRA and MLP parameters are fine-tuned. We fine-tune the model for 50{,}000 iterations using 8 NVIDIA A100 GPUs, with a batch size of 1 per GPU. Training uses the AdamW optimizer with a flat learning rate of $2 \times 10^{-4}$, preceded by a 5{,}000-iteration warmup phase.

\begin{figure}[t]
    \centering
    \begin{subfigure}[b]{0.32\textwidth}
        \centering        \includegraphics[width=\textwidth]{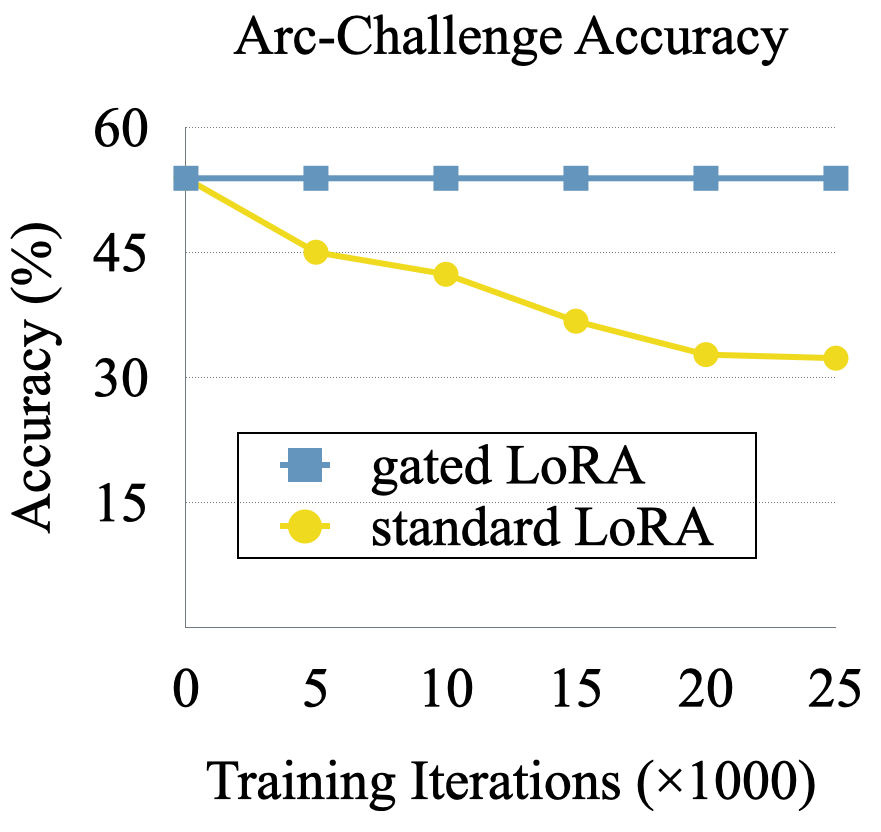}
        \caption{Arc-challenge accuracy}
        \label{fig:arc_challenge}
    \end{subfigure}
    \hfill
    \begin{subfigure}[b]{0.32\textwidth}
        \centering
        \includegraphics[width=\textwidth]{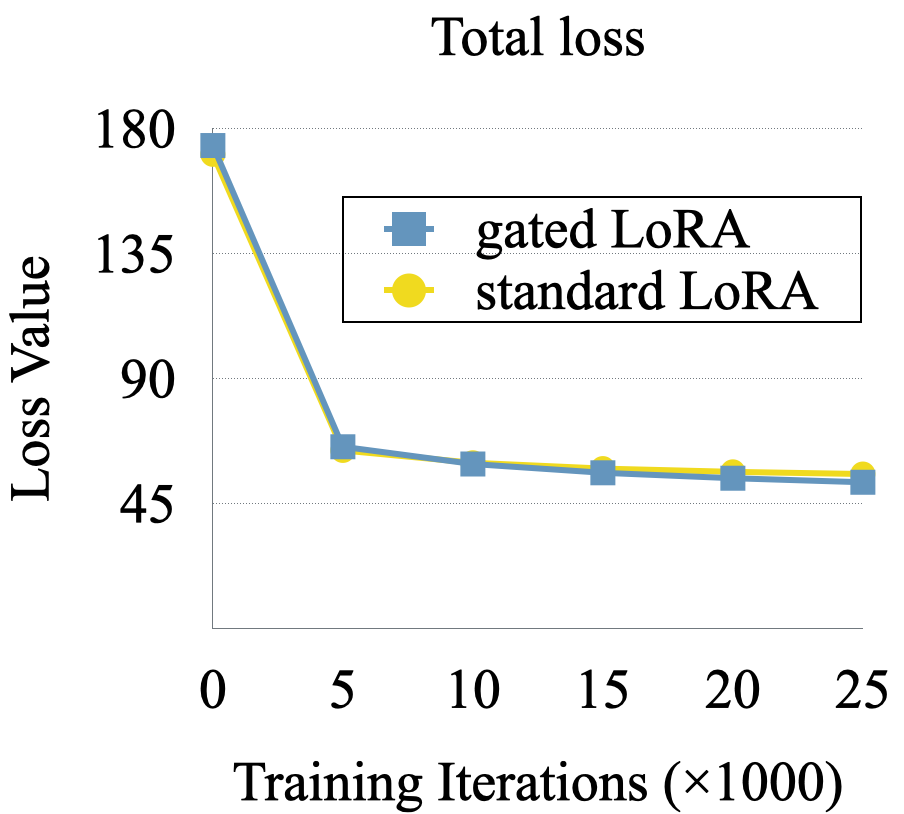}
        \caption{Total loss}
        \label{fig:total_loss}
    \end{subfigure}
    \hfill
    \begin{subfigure}[b]{0.32\textwidth}
        \centering
        \includegraphics[width=\textwidth]{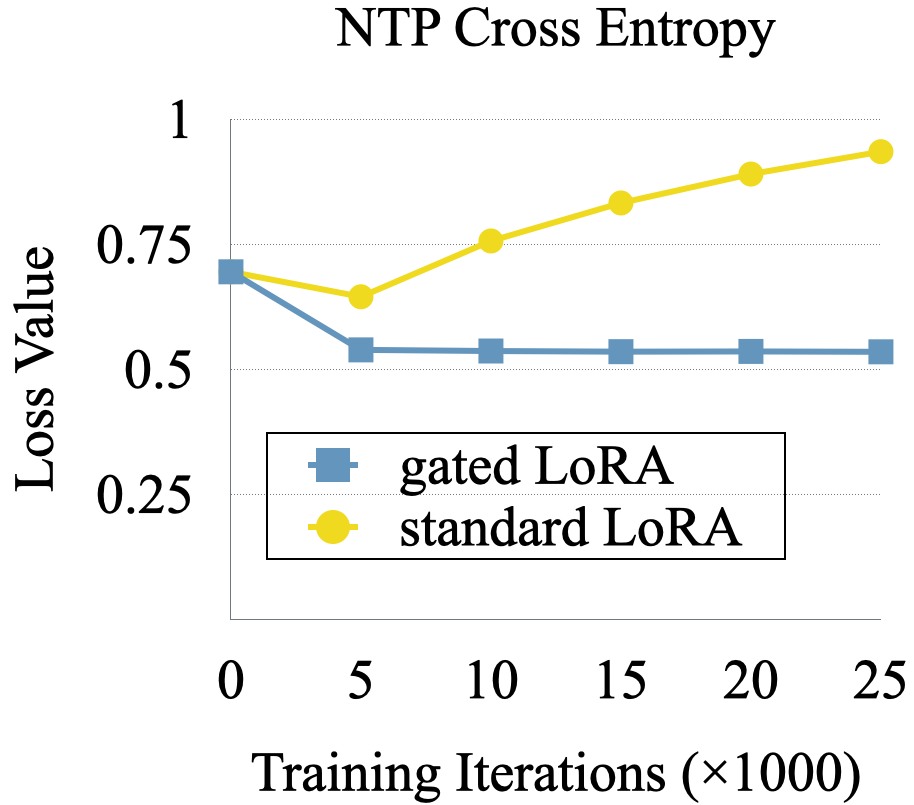}
        \caption{NTP loss}
        \label{fig:ce_0_loss}
    \end{subfigure}
    \caption{Analysis of model convergence trained with  standard and gated LoRA adapters.}
    \label{fig:accuracy_analysis}
    \vspace{-0.5cm}
\end{figure}

\subsection{Generation Quality of Multi Token Predictor}

We begin by tracking model accuracy during fine-tuning. Since we are using an SFT model, it is highly sensitive to additional training. Figure~\ref{fig:arc_challenge} shows the zero-shot accuracy on the ARC-Challenge benchmark, evaluated using the Harness library~\citep{eval-harness}. As shown, the accuracy drops sharply during fine-tuning when the model is augmented with standard LoRA parameters. To address this, we use gated LoRA in place of standard LoRA. As illustrated by the dashed line in Figure~\ref{fig:arc_challenge}, gated LoRA maintains accuracy by ensuring that outputs for NTP tokens remain unchanged.

To investigate the root cause of this degradation, we monitor the total training loss in Figure~\ref{fig:total_loss}. Interestingly, the total loss converges similarly in both the standard and gated LoRA cases. In fact, the loss values are so close that the two curves are nearly indistinguishable. This confirms that both training procedures successfully converge.

Next, we examine the cross-entropy loss over NTP tokens in Figure~\ref{fig:ce_0_loss}, which directly correlates with generation quality. As shown, the model trained with standard LoRA exhibits an increasing trend in NTP loss, indicating degradation. In contrast, the model trained with gated LoRA maintains a nearly constant NTP loss, since gradients do not propagate through NTP tokens when gated LoRA is used.

One might argue that the issues observed with standard LoRA could be mitigated by carefully crafting the training procedure. For example, \citet{cai2024medusa} suggest adjusting MTP loss weights or using separate learning rates. While carefully designed training recipes can partially mitigate the accuracy degradation, they \textit{cannot guarantee} its prevention. Moreover, developing such recipes in practice often involves significant engineering effort and increased GPU costs. In contrast, gated LoRA offers a more robust solution by \textit{ensuring} that the model's original performance remains intact without requiring such manual tuning. The downside of gated LoRA, however, is that the LoRA parameters cannot be fused into the base layer due to the gating function. This downside is a reasonable price to pay for a stable accuracy, especially noting that the overhead of LoRA parameters during inference is reasonably small. Later in our analysis, we show that LoRA parameters with ranks as low as $1$ can effectively learn to predict future tokens.

\subsection{Speedup Analysis}

To assess the speedup achieved by our method, we use the self-speculative decoding algorithms introduced in Section~\ref{sec:specD}. We run up to 100 generation steps, terminating early if the model generates the end-of-sentence token before reaching the 100\textsuperscript{th} step. Given the total number of generation steps $T$ and the total number of tokens generated $G$, we compute the \textit{acceptance rate} as $\frac{G}{T}$. This metric reflects how many tokens are accepted per generation step, indicating the speedup achieved by our method. The acceptance rate has a minimum value of 1, as the model generates at least one token (the next-token prediction) at each step. The maximum theoretical acceptance rate is $k+1=9$, where $k=8$ is the number of masked tokens used during training.

Table~\ref{tab:tulu} reports the acceptance rates of our model across five domains: knowledge, math, coding, chat, and safety. These domains and their associated benchmarks have been verified by the Tulu-3 paper to be free from data contamination relative to the model’s training sets. As shown, our multi-token generation algorithm achieves a speedup ranging from $\approx 1.5$ to $\approx 5.2$, depending on the task and the number of inserted masks. Notably, tasks such as coding and math exhibit higher speedup, likely due to the increased predictability of future tokens in these domains. We also observe diminishing returns in acceptance rate as the number of masks increases: beyond a certain point, additional masks contribute little to further improvement.

\begin{table}[t]
\caption{Speedup obtained by multi-token-prediction over different domains for Tulu-3.}\label{tab:tulu}
\resizebox{\textwidth}{!}{
\begin{tabular}{cccccccccc}
\multirow{2}{*}{\textbf{Skill}}     & \multirow{2}{*}{\textbf{Benchmark}} & \multicolumn{8}{c}{\textbf{Acceptance Rate}}                                                                                                         \\
                                    &                                     & \textbf{1 mask} & \textbf{2 masks} & \textbf{3 masks} & \textbf{4 masks} & \textbf{5 masks} & \textbf{6 masks} & \textbf{7 masks} & \textbf{8 masks} \\ \hline
\multirow{3}{*}{\textbf{Knowledge}} & \textbf{MMLU}                       & 1.54            & 1.87             & 2.06             & 2.18             & 2.26             & 2.32             & 2.36             & 2.38             \\
                                    & \textbf{PopQA}                      & 1.48            & 1.69             & 1.80             & 1.85             & 1.88             & 1.90             & 1.91             & 1.91             \\
                                    & \textbf{TruthfulQA}                 & 1.55            & 1.84             & 2.00             & 2.08             & 2.13             & 2.16             & 2.18             & 2.19             \\ \hline
\textbf{Math}                       & \textbf{GSM8k}                      & 1.84            & 2.58             & 3.21             & 3.75             & 4.20             & 4.60             & 4.95             & 5.22             \\ \hline
\textbf{Coding}                     & \textbf{HumanEval}                  & 1.86            & 2.63             & 3.30             & 3.87             & 4.40             & 4.79             & 5.14             & 5.35             \\ \hline
\multirow{2}{*}{\textbf{Chat}}      & \textbf{AlpacaEval}                 & 1.61            & 1.97             & 2.18             & 2.31             & 2.40             & 2.45             & 2.49             & 2.52             \\
                                    & \textbf{IFEval}                     & 1.55            & 1.86             & 2.05             & 2.16             & 2.24             & 2.30             & 2.36             & 2.39             \\ \hline
\multirow{6}{*}{\textbf{Safety}}    & \textbf{XSTest}                     & 1.70            & 2.11             & 2.38             & 2.52             & 2.67             & 2.73             & 2.76             & 2.79             \\
                                    & \textbf{HarmBench}                  & 1.78            & 2.35             & 2.76             & 2.99             & 3.28             & 3.39             & 3.45             & 3.51             \\
                                    & \textbf{Do-Anything-Now}            & 1.69            & 2.29             & 2.85             & 3.04             & 3.21             & 3.23             & 3.64             & 3.72             \\
                                    & \textbf{JailbreakTrigger}           & 1.77            & 2.31             & 2.76             & 3.02             & 3.31             & 3.41             & 3.48             & 3.54             \\
                                    & \textbf{WildJailbreakTest}          & 1.71            & 2.18             & 2.47             & 2.64             & 2.79             & 2.84             & 2.88             & 2.91             \\
                                    & \textbf{WildGuardTest}              & 1.69            & 2.12             & 2.39             & 2.54             & 2.67             & 2.72             & 2.76             & 2.79             \\ \hline
\multicolumn{2}{c}{\textbf{Average}}                                      & 1.67            & 2.14             & 2.48             & 2.69             & 2.88             & 2.99             & 3.10             & 3.17            
\end{tabular}
}
\end{table}

\begin{figure}[ht]
    \centering
    \includegraphics[width=0.75\linewidth]{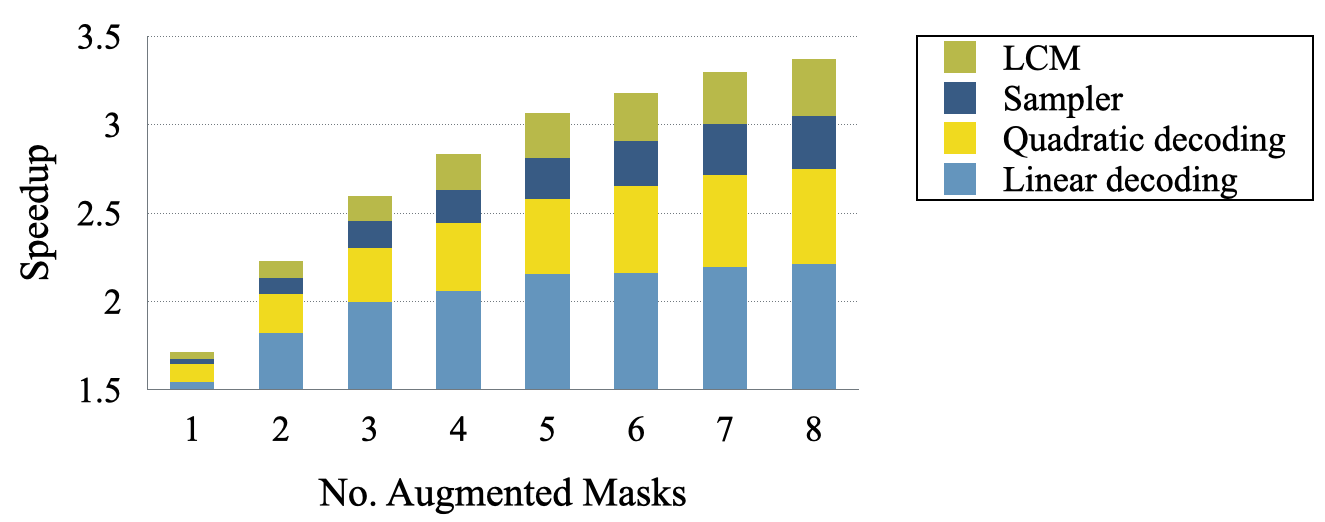}
    \caption{Average speedup achieved by the simplest and most advanced configurations of our MTP models. The basic version uses linear decoding without LCM loss or a sampler head, while the advanced version includes LCM loss, a sampler head, and quadratic decoding.}
    \label{fig:first_vs_last}
    \vspace{-0.5cm}
\end{figure}

\subsection{Ablation Studies}

The results in Table~\ref{tab:tulu} correspond to the best-performing configuration of our method, which includes: (1) the use of a sampler MLP head, (2) the application of LCM loss during training, and (3) the use of a quadratic decoding algorithm during generation. We will ablate the contribution of each component hereafter. In Figure~\ref{fig:first_vs_last}, we compare the impact of each of the components introduced in this paper on the speedup. The most basic version is shown in light blue, which has no sampler head, no LCM loss during training, and utilizes a linear decoding algorithm. The next improvement (yellow) is to apply quadratic decoding rather than linear decoding, which shows higher acceptance rates. Notably, the improvement is more pronounced when a larger number of masks is used. This is because, with more inserted masks, the likelihood of future tokens being rejected increases under the linear decoding strategy, providing more opportunity for the quadratic decoding algorithm to outperform it.

In the next experiment of Figure~\ref{fig:first_vs_last}, we evaluate the impact of using the sampler MLP head on speedup (dark blue). The results indicate that incorporating the sampler head leads to improved speedup, with the effect becoming more pronounced as the number of masked tokens increases. This is due to the fact that achieving coherency in one shot is extremely challenging when the sequence length is long, which necessitates the use of the sampler head to address this challenge.

Finally, we assess the impact of the LCM loss (olive green), also shown in Figure~\ref{fig:first_vs_last}. The results demonstrate that the LCM loss improves the alignment between the model’s predictions and those of the underlying autoregressive model, ultimately leading to higher overall speedup.

\begin{figure}[h]
    \centering
    \includegraphics[width=1\linewidth]{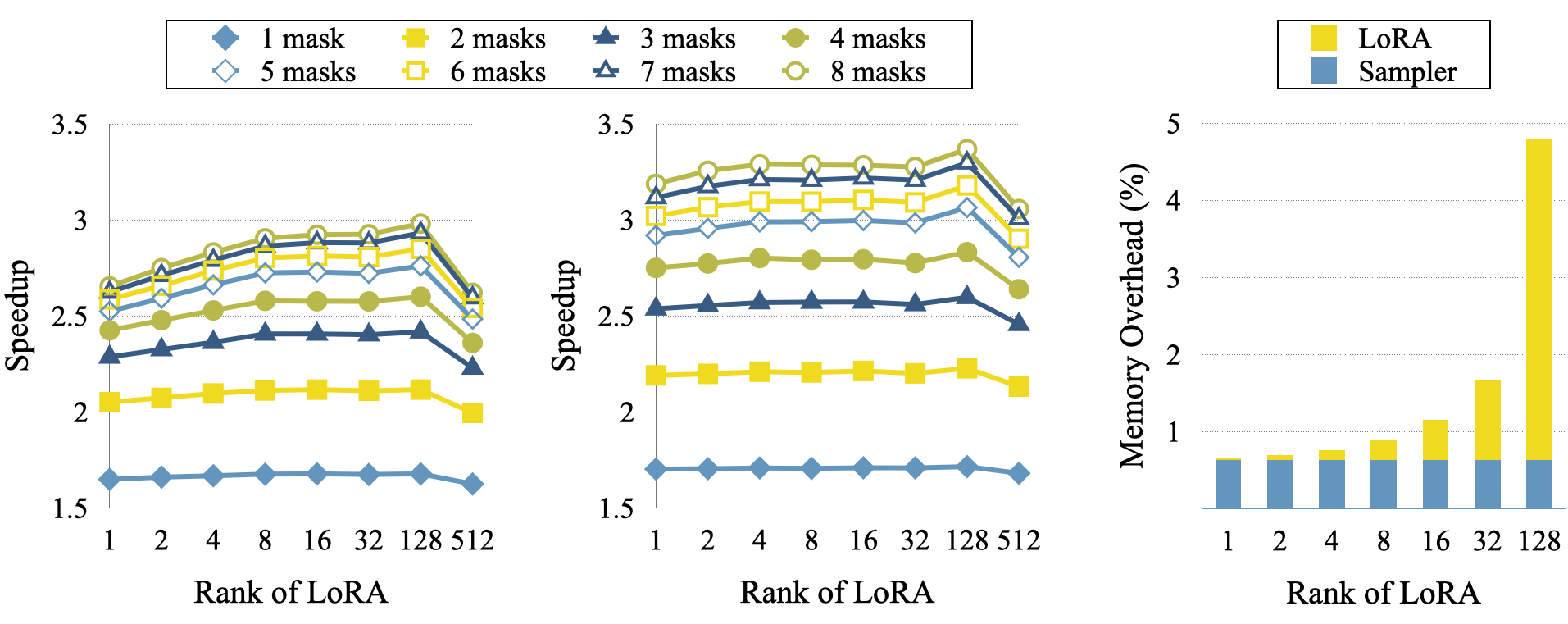}
    \caption{Effect of LoRA rank: Speedup without the sampler head (left), speedup with the sampler head (middle), and memory overhead of LoRA and the sampler head (right).}
    \label{fig:lora}
\end{figure}

The experiments discussed so far used a model augmented with LoRA parameters of rank 128. In the following analysis, we train several models with varying LoRA ranks. Figure~\ref{fig:lora} summarizes the results: the left and middle panels show acceptance rates without and with the sampler head, respectively, while the right panel shows the memory overhead of both the sampler head and the LoRA parameters. We make three main observations in these experiments. 

First, the MTP model without the sampler head (left panel) maintains reasonable speedup even with very low ranks—such as 16, 4, or even 1. Mapping these ranks to the right panel, we see that the memory overhead of LoRA at such small ranks is negligible. This supports our main hypothesis: the pretrained NTP model already contains substantial knowledge about future tokens, and even a small-rank augmentation can organize and leverage this knowledge to achieve MTP speedup.

Second, comparing the left and middle panels, we observe that adding the sampler head has a more significant impact on acceptance rate than simply increasing the LoRA rank.

Third, increasing the rank beyond 128 negatively impacts the speedup. While the exact cause remains unclear, one possible explanation is that training the model on a relatively small SFT dataset leads to overfitting when an unnecessarily large number of trainable parameters is introduced.

\section{Related Works}
\textbf{Extension of autoregressive models to MTP:} A key contribution of our work is the insight that autoregressive models are inherently capable of predicting future tokens. We show that this ability can be unlocked with minimal fine-tuning and a small number of additional parameters. This approach retains the strengths of autoregressive models while also enabling the efficiency gains of multi-token prediction.

\textbf{Self speculation for speedup:} To evaluate the speedup gained by our MTP method, we consider the self-speculation framework for speculative decoding ~\citep{Leviathan2022FastIF}, which has been used successfully with other MTP-based models \citep{cai2024medusa} and EAGLE \citep{Li2024EAGLESS,Li2024EAGLE2FI,Li2025EAGLE3SU}. To further improve the speedup achieved by our method, we introduce a quadratic decoding algorithm. Later in our research, we discovered that the tree-based exploration algorithm proposed by~\citep{Chen2024HardwareAwarePP} is compatible with our approach and can be integrated with our quadratic decoding.

\textbf{Harnessing the full LLM instead of adding heads:} Existing MTP methods have made significant progress by effectively reducing the number of inference calls, marking an important milestone in improving generation efficiency. A common strategy among these approaches is the use of additional prediction heads \citep{cai2024medusa}, \citep{Li2024EAGLESS}, \citep{liu2024deepseek}, which, while effective, can increase both inference cost and parameter count. In contrast, our method takes a different path by augmenting the input sequence to prompt the pretrained model to generate multiple token predictions. This allows us to parallelize the compute used by the existing model architecture without introducing substantial latency or additional components.

\textbf{Using mask tokens for MTP:} 
Mask tokens have been explored for multi-token prediction (MTP) in general \citep{Monea2023PaSSPS}, and more specifically in speculative decoding \citep{Gerontopoulos2025MultiTokenPN,Chen2024HardwareAwarePP,Liu2024SDSATAL,Xiao2024ParallelSpecPD}. These works have demonstrated the effectiveness of mask-based strategies for accelerating generation and have laid important groundwork for further improvements in MTP using masks. These approaches typically fall into two categories: either fine-tuning the entire model, which sacrifices the accuracy of next-token prediction, or only adjusting a small number of parameters in the token embedding table, which limits flexibility. In contrast, we propose gated LoRA, a method that preserves the original autoregressive performance while allowing selective fine-tuning. Gated LoRA enables us to fine-tune all layers of the model, yet introduces only minimal parameter and computational overhead. While prior work has explored soft gating \citep{Liang2025GatedIO} and dropout-based training of LoRA \citep{Wang2024MLAEML}, our approach is, to the best of our knowledge, the first to use fixed binary masking of LoRA parameters. This provides a simple and reliable mechanism to preserve original model accuracy while still allowing finetuning for behavior on some tokens.

\textbf{Lightweight sampling at inference:}
A key contribution of our method is the introduction of a lightweight, MLP-based, sampler head with minimal parameters. This trained module serves as a replacement for traditional beam search \citep{cheng2024recurrent}, avoiding its complexity and runtime cost. While inspired by prior work \citep{liu2024deepseek}, our sampler differs in a crucial way: it does not perform multi-token prediction itself. Instead, the main model generates a rich distribution over future tokens, and the sampler head selects a coherent token sequence from this distribution. This separation allows our sampler to remain highly efficient, requiring only a two-layer MLP rather than multiple transformer blocks as in \citep{liu2024deepseek}. Nevertheless, our method can be combined with beam search or more complex recurrent sampling heads, such as the one used in \citep{liu2024deepseek}.

This is similar to \citep{Ankner2024HydraSD} which similarly proposes the use of MLPs to condition on past tokens in the speculation, however, in their case, the input size of each MLP grows linearly to take all previous speculative tokens as input, whereas we condition only on the most recently decoded speculative token, keeping the MLP size constant.

\textbf{Latent consistency loss}: We propose to use a consistency loss to improve the accuracy of future token prediction. Similarly to our work on consistency loss, knowledge distillation \citep{Hinton2015DistillingTK, Kim2016SequenceLevelKD} is widely used in many multi-token prediction and speculative decoding approaches \citep{Zhou2023DistillSpecIS, Chen2024HardwareAwarePP, cai2024medusa, Li2024EAGLESS}. Knowledge distillation and consistency loss can serve both as a means of aligning the speculative drafter with the verification model, and as a mechanism for reducing output modalities—enabling the non-autoregressive draft model to learn and predict more effectively \citep{Gu2017NonAutoregressiveNM, Zhou2019UnderstandingKD}. Our specific formulation of consistency loss was inspired by its use in distillation \citep{Luo2023LatentCM}.
\section{Conclusion}
We start from the observation that vanilla, autoregressive, language models contain large amounts of information about future tokens beyond the immediate next token. We then explored and combined various techniques to materialize the potential of autoregressive models to generate multiple tokens.
We craft a training recipe with mask tokens which can better the prediction accuracy for future tokens, while maintaining the performance of the immediate next token generation. Using speculative decoding, we provide an important measure by which we can evaluate the amount of usable information (beyond simple, token-wise independent metrics, perplexity, etc.). Using this metric, we show that additional improvements to the modeling of future tokens can be made with lightweight additions to the base model that do not compromise its autoregressive generation performance.

We have evaluated how well autoregressive models can adapt to multi-token prediction during the supervised fine-tuning stage of language model training. A promising direction for future work is to explore the impact of this approach during pretraining or downstream task adaptation. Another compelling avenue is to investigate diffusion-based generation for multi-token prediction. We believe that multi-token prediction lies between fully autoregressive and fully diffusion-based generation methods, offering a balanced combination of advantages from both ends of the spectrum.


\bibliographystyle{unsrtnat}
\bibliography{references}









\end{document}